%% file: main.tex
\documentclass[10pt,twocolumn,letterpaper]{article}

\usepackage{cvpr}              

\input{preamble}

\definecolor{cvprblue}{rgb}{0.21,0.49,0.74}
\usepackage[pagebackref,breaklinks,colorlinks,allcolors=cvprblue]{hyperref}

\def\paperID{9894} 
\def\confName{CVPR}
\def\confYear{2026}

\title{
FlashVGGT: Efficient and Scalable Visual Geometry Transformers with Compressed Descriptor Attention
}

\author{Zipeng Wang \quad Dan Xu\thanks{Corresponding author}\\
The Hong Kong University of Science and Technology\\
{\tt\small zwang253@connect.ust.hk} \quad {\tt\small danxu@cse.ust.hk}
}

\begin{document}

\input{figures/figure1/figure1}

\input{sec/0-abstract}    

\input{sec/1-introduction}

\input{sec/2-related-work}

\input{figures/figure3/figure3}
\input{sec/3-method}

\input{sec/4-experiment}

\input{sec/5-conclusion}

\input{sec/6-acknowledgment}

{
    \small
    \bibliographystyle{ieeenat_fullname}
    \bibliography{main}
}

\input{sec/X_suppl}

\end{document}

%% file: preamble.tex
\renewcommand{\paragraph}[1]{\vspace{.5em}\noindent\textbf{#1.}}

\usepackage{booktabs}
\usepackage{multirow}
\usepackage{graphicx}
\usepackage{siunitx}
\usepackage[normalem]{ulem}
\useunder{\uline}{\ul}{}
\usepackage{adjustbox}

%% file: figures/figure1/figure1.tex
\twocolumn[{%
\renewcommand\twocolumn[1][]{#1}%
\maketitle
\includegraphics[width=\textwidth]{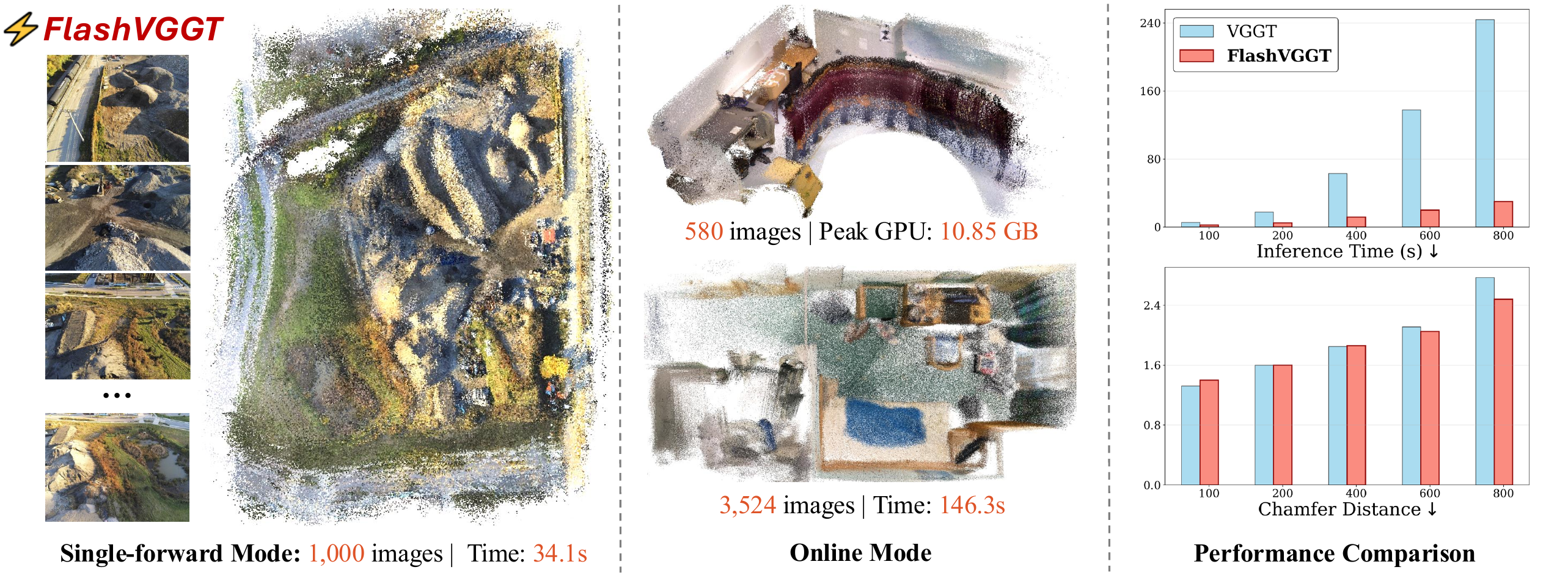}
\captionof{figure}{
\small \textbf{FlashVGGT} achieves significant speedup and scales to larger inputs. 
Our method enables both fast single-forward inference on long sequences (left) and memory-efficient online inference (center)  while maintaining competitive accuracy versus VGGT~\cite{wang2025vggt} (right).
\vspace{1.5em}}
\label{fig:teaser}
}]

%% file: sec/0-abstract.tex
\begin{abstract}
3D reconstruction from multi-view images is a core challenge in computer vision. Recently, feed-forward methods have emerged as efficient and robust alternatives to traditional per-scene optimization techniques. Among them, state-of-the-art models like the Visual Geometry Grounding Transformer (VGGT) leverage full self-attention over all image tokens to capture global relationships. However, this approach suffers from poor scalability due to the quadratic complexity of self-attention and the large number of tokens generated in long image sequences.
In this work, we introduce FlashVGGT, an efficient alternative that addresses this bottleneck through a descriptor-based attention mechanism. Instead of applying dense global attention across all tokens, FlashVGGT compresses spatial information from each frame into a compact set of \textbf{descriptor tokens}. 
Global attention is then computed as cross-attention between the full set of image tokens and this smaller descriptor set, significantly reducing computational overhead.
Moreover, the compactness of the descriptors enables online inference over long sequences via a chunk-recursive mechanism that reuses cached descriptors from previous chunks. Experimental results show that FlashVGGT achieves reconstruction accuracy competitive with VGGT while reducing inference time to just 9.3\% of VGGT for 1,000 images, and scaling efficiently to sequences exceeding 3,000 images.
Our project page is available at \texttt{\url{https://wzpscott.github.io/flashvggt_page/}}.

\end{abstract}

%% file: sec/1-introduction.tex
\section{Introduction}
Reconstructing 3D geometry from multi-view images is a fundamental problem in computer vision~\cite{hartley2003mvgbook}.
Traditional pipelines such as Structure-from-Motion (SfM)~\cite{schoenberger2016sfm, ullman1979interpretation, koenderink1991affine, pan2024glomap} and Multi-View Stereo (MVS)~\cite{seitz2006mvs, schoenberger2016mvs, moulon2016openmvg} have dominated this task for decades.
These methods rely on per-scene, iterative optimization pipelines that include feature detection, matching, triangulation, and bundle adjustment.
While often accurate, these pipelines are computationally intensive and fragile, requiring extensive processing and careful tuning for each scene, especially under challenging conditions~\cite{lowe2004distinctive, sarlin2020superglue}.

Recent advances have shifted toward learning-based approaches that bypass the complexities of traditional pipelines by directly predicting 3D structure using neural networks~\cite{wang2024dust3r, leroy2024mast3r, yang2025fast3r, wang2024vggsfm}.
These models are trained end-to-end on large-scale datasets~\cite{reizenstein2021co3d, yao2020blendedmvs, MVSSynth, neuhold2017mapillary, baruch2021arkitscenes, dai2017scannet}, enabling direct 3D prediction from multi-view input.
This feed-forward paradigm eliminates the need for sequential post-processing and per-scene optimization.
Moreover, they exhibit improved robustness by learning strong priors from diverse, large-scale data.

\input{figures/figure2/figure2}

A recent milestone in this direction is the Visual Geometry Grounding Transformer (VGGT)~\cite{wang2025vggt}, which performs high-fidelity 3D reconstruction from hundreds of views in a single forward pass.
VGGT’s success stems from an alternating attention backbone that combines frame-wise and global attention blocks~\cite{vaswani2017attention, dosovitskiy2020vit}, enabling effective aggregation of frame and global context.
However, this architecture has a key limitation: the global attention block requires self-attention over all image tokens.
As shown in Fig.~\ref{fig:figure2(a)}, this leads to quadratic complexity, creating a severe bottleneck and limiting scalability of VGGT to long sequences.

This work is driven by a central question: \textit{Is full self-attention truly necessary for global reasoning in VGGT?}
We base our approach on two key insights.
First, classical methods show that accurate inter-frame associations can be inferred from sparse keypoints and descriptors~\cite{lowe1999sift, bay2008surf}, suggesting that dense token-to-token attention may be unnecessary.
Second, we observe that VGGT’s global attention maps are inherently sparse, with most scores near zero (Fig.~\ref{fig:figure2(b)}), implying that much of the computation is spent on irrelevant token pairs.
These observations motivate our quest for a more efficient alternative that retains global reasoning while scaling to long input sequences.

Motivated by these insights, we propose \textbf{FlashVGGT}, an efficient architecture that overcomes the computation bottleneck of VGGT through compressed descriptor attentions. 
Our core innovation reformulates the global attention block by generating a compact set of descriptor tokens via spatial resampling, which encapsulate key information from each frame. 
Global attention is then approximated via cross-attention from image tokens to descriptors (\ie, using image tokens as queries and descriptors as keys/values). 
This design reduces the computational complexity of global attention from $\mathcal{O}(N^2)$ to $\mathcal{O}(N^2 / r^2)$, where $r$ is the spatial compression ratio. 
Empirically, FlashVGGT achieves over 90\% inference speedup on 1,000-image sequences, with accuracy comparable to VGGT.

Furthermore, the compactness of our descriptor tokens enables scalable inference for very long sequences (e.g., 3,000 images) through an chunk-recursive scheme. 
When processing sequences that exceed memory limits, we divide the input into sequential chunks. By caching and reusing descriptor tokens from previous chunks, later chunks incorporate historical context while maintaining a global receptive field across the entire sequence. Crucially, unlike StreamVGGT~\cite{zhuo2025streamvggt}, which caches full-resolution tokens from all transformer layers, our method only stores the compressed descriptors. This achieves an $r^2$ reduction in peak memory usage, enabling scalable reconstruction for substantially larger inputs and resource-constrained scenarios.

The main contributions of this work are summarized as follows:
\textbf{(i)} We propose FlashVGGT, an efficient framework that alleviates the quadratic complexity of global attention in VGGT.
\textbf{(ii)} We design a chunk-recursive inference mechanism that enables online reconstruction of long sequences using cached descriptors.
\textbf{(iii)} Extensive experiments demonstrate that FlashVGGT achieves competitive accuracy while reducing inference time by over 90\% on 1,000-image inputs and scales to over 3,000 views.

%% file: figures/figure2/figure2.tex
\begin{figure}[tbp]
    \centering
    
    \begin{subfigure}{\linewidth}
        \centering
        \includegraphics[width=.95\linewidth]{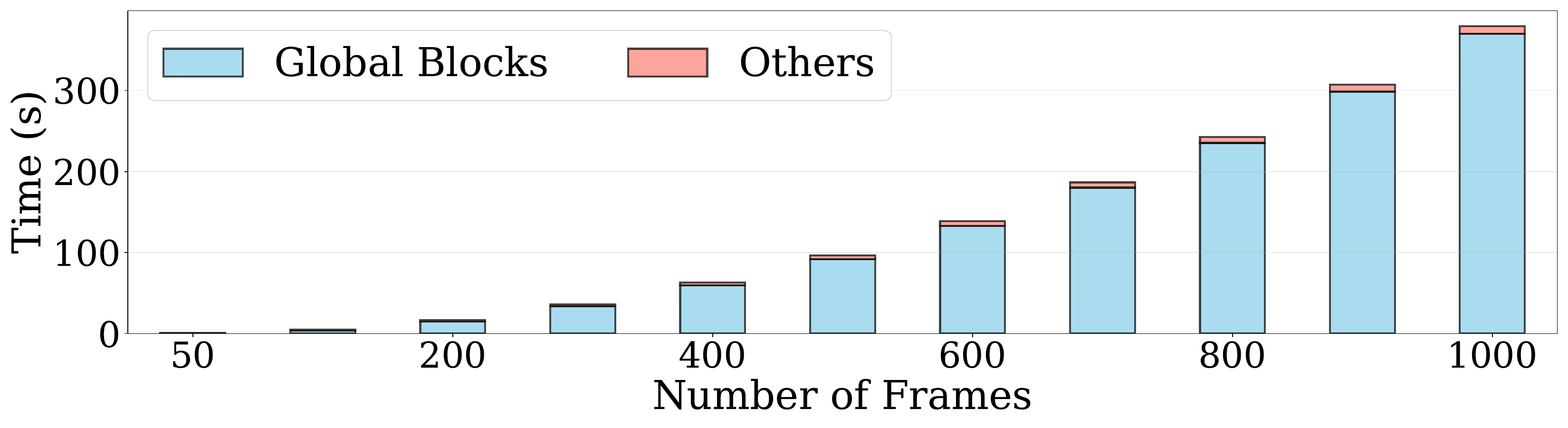}
        \caption{Computational Breakdown of VGGT}
        \label{fig:figure2(a)}
    \end{subfigure}
    
    \vspace{1pt} 
    
    \begin{subfigure}{\linewidth}
        \centering
        \includegraphics[width=.95\linewidth]{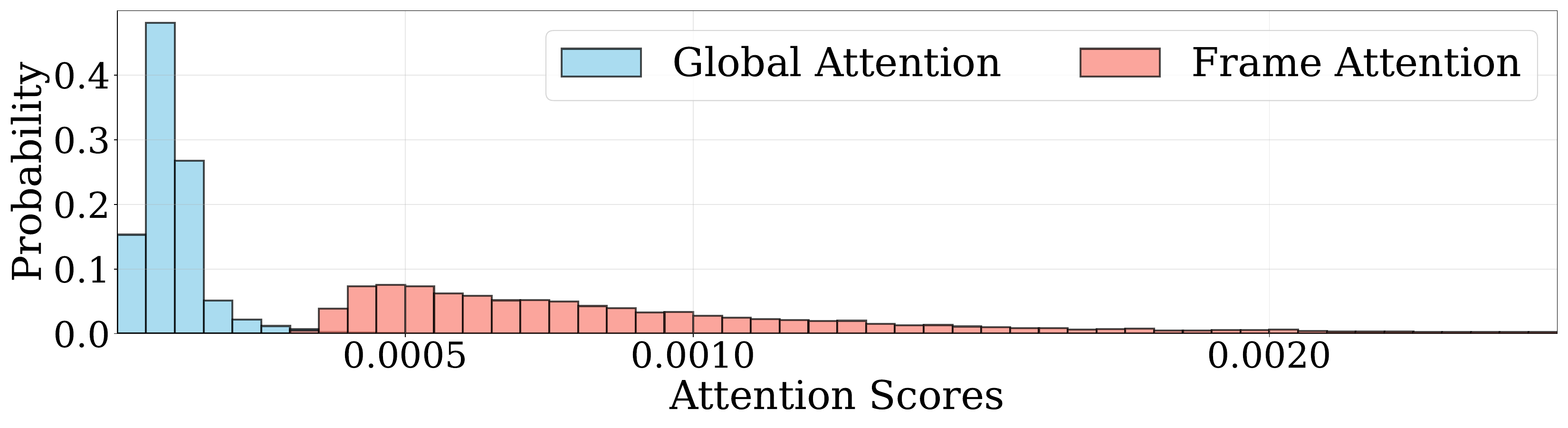}
        \caption{Histogram of Attention Scores Distributions}
        \label{fig:figure2(b)}
    \end{subfigure}
    
    \caption{
    \textbf{(a) }The global attention block is the primary computational bottleneck in VGGT, dominating total inference time.
    \textbf{(b) }Global attention is highly sparse, with most scores concentrated near zero, suggesting that full self-attention is highly inefficient. In contrast, frame attention exhibits a more uniform distribution.
    }
    \label{fig:custom_stacked}
\end{figure}

%% file: sec/2-related-work.tex
\section{Related Work}

\noindent \textbf{Feed-forward 3D Reconstruction.}
An emerging paradigm in 3D reconstruction is to directly predict 3D structures from images using deep-learning models trained on large datasets. This paradigm, often referred to as feed-forward 3D reconstruction, replaces per-scene iterative optimization with one or several forward passes of a neural network, offering greater efficiency and robustness than traditional methods.
Early efforts in this line of research~\cite{wang2024dust3r, leroy2024mast3r, zhang2024monst3r} predict pairwise 3D point maps and employ per-scene global alignment to reconstruct from multiple views. 
Subsequent works~\cite{yang2025fast3r, wang2025pi3, keetha2025mapanything} sought to directly predict 3D structures from multiple images in a single forward pass.
Notably, VGGT~\cite{wang2025vggt} introduced a transformer-based architecture with alternating frame-wise and global attention blocks. This approach enables reconstruction from hundreds of images in a single forward pass with high accuracy. However, VGGT's global attention module requires full self-attention over all image tokens, causing computational cost to scale nearly quadratically with the number of input images. This results in significant overhead for large-scale inputs. 
Another line of research~\cite{wang2025cut3r, chen2025ttt3r, deng2025vggtlong} performs 3D reconstruction in an online manner. While these methods offer improved memory efficiency, their reconstruction accuracy often lags behind that of offline counterparts..

\noindent \textbf{Efficient Vision Transformers.}
Vision Transformers~\cite{dosovitskiy2020vit, liu2021swintransformer, arnab2021vivit} have become a pivotal component in modern computer vision, powered by the self-attention mechanism. However, a major limitation of Vision Transformers is the quadratic computational complexity of self-attention~\cite{dao2022flashattention, xiao2023streamingllm,tang2024quest,li2025radial}, which leads to prohibitive overhead for large-scale inputs.
Many efforts~\cite{vasu2023fastvit, liu2023efficientvit, li2022efficientformer, dao2022flashattention} have been made to mitigate the computational demands of self-attention. One line of research introduces sparsity~\cite{chen2021chasingsparsity, wei2023sparsifiner} or low-rank approximations~\cite{jaegle2021perceiver, han2024agent} into the attention computation. Another direction focuses on reducing the number of tokens involved in the attention operation by clustering~\cite{marin2021tokenpooling}, merging\cite{bolya2022tokenmerging, bolya2023tokenmergingdiffusion} or selection~\cite{wang2022efficientselection, fayyaz2022adaptiveselection}.
Despite these advancements, most existing efficient Transformer methods are designed for 2D tasks like image classification. 
Consequently, it remains an open challenge to design efficient Vision Transformers for 3D reconstruction, as it demands maintaining long-range, multi-view geometric consistency.

\noindent \textbf{Concurrent Work.}
Concurrently with our work, several other approaches have been proposed to address the efficiency bottleneck in VGGT. 
FastVGGT~\cite{shen2025fastvggt} employs token merging to reduce the number of tokens fed into global attention. However, the process of identifying and merging similar tokens across the entire sequence introduces considerable computational overhead. 
FasterVGGT~\cite{wang2025fastervggt} introduces block sparsity into the attention matrix to reduce computation. While effective for moderate sparsity levels, this approach suffers from significant performance degradation when sparsity increases.
StreamVGGT~\cite{zhuo2025streamvggt} enables incremental reconstruction from streaming input by caching intermediate tokens from all global attention blocks. This design, however, leads to substantial memory overhead, limiting its scalability to longer sequences beyond tens of images.
In contrast, FlashVGGT provides a more principled approach to global context compression, enabling a superior balance among inference speed, memory consumption and accuracy compared to these concurrent approaches.


%% file: figures/figure3/figure3.tex
\begin{figure*}[!t]
    \centering
    \includegraphics[width=0.99\linewidth]{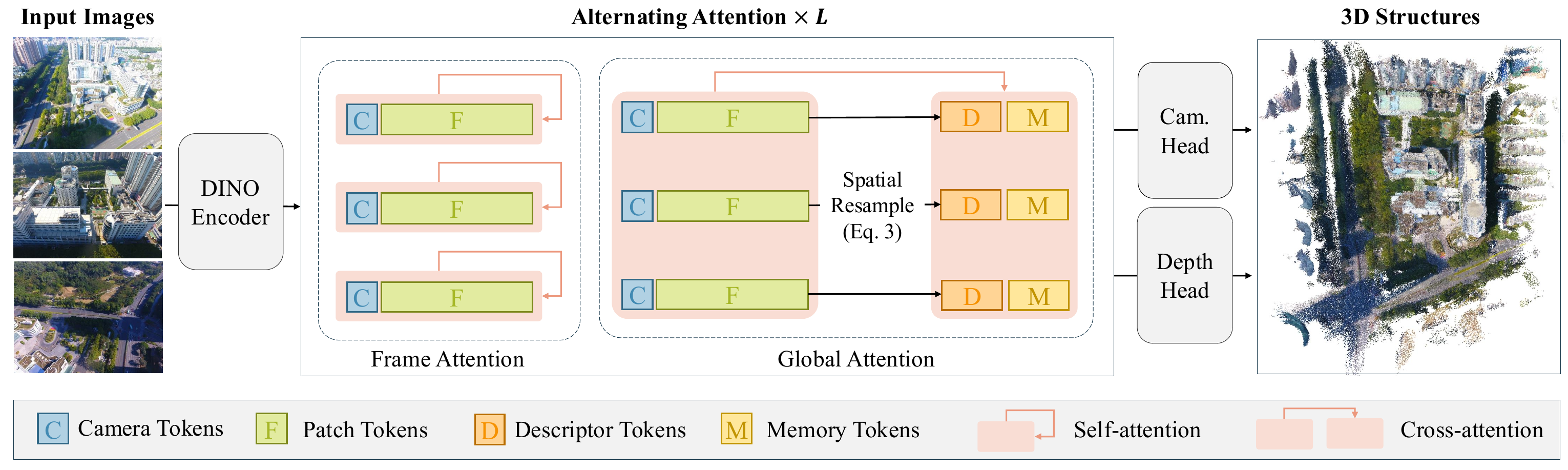}
    \caption{
    \textbf{Architecture Overview.}
    Our framework encodes input images into tokens using DINO~\cite{caron2021dino} and processes them through alternating frame and global attention blocks. Unlike VGGT's dense global attention over all tokens, FlashVGGT generates a compact set of descriptor tokens via spatial compression and computes efficient cross-attention from image tokens to these descriptors. The final aggregated tokens are fed into reconstruction heads to predict camera parameters and depth maps.
    }
    \label{fig:figure3}
\end{figure*}

%% file: sec/3-method.tex
\section{Method}
\subsection{Visual Geometry Grounding Transformers}
\label{ssec:vggt}

We build our approach upon the Visual Geometry Grounding Transformer (VGGT)~\cite{wang2025vggt}. For completeness, we briefly recap its architecture, which consists of three main stages: image encoding, feature aggregation via alternating attention, and 3D reconstruction heads.

\noindent \textbf{Image Encoder.} 
Given a sequence of $S$ input images $\{I_i\}_{i=1}^S$, a DINO~\cite{caron2021dino} encoder extracts a set of feature tokens for each image. This results in a set of token sequences $\mathbf{F} = \{\mathbf{F}_1, \mathbf{F}_2, \dots, \mathbf{F}_S\}$, where each $\mathbf{F}_i \in \mathbb{R}^{N \times C}$ represents the $N$ tokens from the $i$-th image. Each sequence includes a special learnable camera token that stores camera information and several register tokens~\cite{darcet2023register}.

\noindent \textbf{Alternating Attention.} The core of VGGT is a transformer aggregator composed of $L$ identical layers, each containing two attention blocks designed to capture both intra- and inter-frame relationships. For simplicity, we omit the layer index in the following formulations.
\begin{itemize}
    \item \textbf{Frame Attention} processes tokens \textit{within} each frame independently. For each frame $i$, self-attention is computed over its $N$ tokens to refine local features:
    \begin{equation}
        \mathbf{F}_i' = \text{SelfAttn}(\mathbf{F}_i), \quad \mathbf{F}_i \in \mathbb{R}^{N \times C}
    \end{equation}
    \item \textbf{Global Attention} models interactions \textit{across all frames}. The tokens are concatenated into a single global sequence $\mathbf{G} \in \mathbb{R}^{K \times C}$, where $K = S \times N$. Standard self-attention is then computed over all $K$ tokens:
    \begin{equation}
        \mathbf{H} = \text{SelfAttn}(\mathbf{G})
        \label{eq:original-global-attn}
    \end{equation}
    The output $\mathbf{H} \in \mathbb{R}^{K \times C}$ is then reshaped back into $S$ frame sequences for subsequent processing.
\end{itemize}

\noindent \textbf{Reconstruction Heads.} 
The final aggregated token features are used by two separate heads to predict 3D properties for each view: (i) a \textbf{camera head} predicts camera extrinsics and intrinsics, and (ii) a \textbf{DPT head}~\cite{ranftl2021dpt} predicts a depth map and an aleatoric uncertainty map~\cite{kendall2016uncertainty}.

While this architecture is powerful, its scalability is limited by the quadratic complexity $\mathcal{O}(K^2) = \mathcal{O}(S^2N^2)$ of the dense global self-attention, as identified in Fig.~\ref{fig:figure2(a)}.

\subsection{Descriptor-Based Global Attention}
\label{ssec:efficient_attn}

We introduce an efficient alternative to the dense global attention block that preserves global reasoning while reducing its complexity. Our approach replaces standard self-attention in global attention blocks with a descriptor-based cross-attention mechanism, while keeping the encoder, frame attention, and reconstruction heads unchanged due to their minimal computational overhead. Fig.~\ref{fig:figure3} provides an overview of our framework.

\noindent \textbf{Spatially-Compressed Descriptor Tokens.} 
Given the input to a global block, $\mathbf{G} = \text{Reshape}(\{\mathbf{F}_i'\}_{i=1}^S) \in \mathbb{R}^{K \times C}$, we first restore its spatial structure by reshaping it to $\mathbf{G} \in \mathbb{R}^{S \times H \times W \times C}$, where $H$ and $W$ are the height and width of the 2D patch token grid for each frame.
We then generate a compact set of descriptor tokens $\mathbf{D}$ by applying spatial compression. 
Compared to pooling-based methods, interpolation better preserves local spatial information in the original features (see discussion in Sec.~\ref{compress_method}). We employ bilinear interpolation to resample each frame's spatial dimensions $(H, W)$ to a lower resolution $\left(\lfloor H/r \rfloor, \lfloor W/r \rfloor\right)$, where $r$ is the compression factor:
\begin{equation}
    \mathbf{D} = \text{Reshape}\left(\text{Interp}\left(\mathbf{G}, \left(\lfloor H/r \rfloor, \lfloor W/r \rfloor\right)\right)\right)
\end{equation}
The resulting descriptor tokens have a size of $\mathbb{R}^{K_d \times C}$, where $K_d = S \times \lfloor H/r \rfloor \times \lfloor W/r \rfloor$. 

\noindent \textbf{Auxiliary Descriptor Tokens.} \label{aux_tokens}
To maintain geometric consistency, we augment the compressed descriptors with three types of auxiliary tokens: 
(i) camera and register tokens from all frames; 
(ii) all tokens from the first image (which defines the world coordinate system)
and (iii) all tokens from key-frames selected via k-means clustering~\cite{lloyd1982kmeans} on average frame tokens. The key-frame selection is highly efficient, converging in under 2 seconds for 1,000 images on a single NVIDIA H800 GPU as it operates on per-frame averages rather than individual tokens.
These auxiliary tokens act as geometric anchors, preserving high-fidelity information from camera parameters, the world coordinate frame, and representative views. This prevents the loss of critical details during descriptor compression, ensuring robust geometric reasoning across the entire sequence.

\noindent \textbf{Descriptor Attention.} 
We reformulate the global attention operation from Eq.~(2) as a cross-attention layer. The original, full-resolution tokens $\mathbf{G}$ are used as queries, while the descriptor tokens $\mathbf{D}$ are used as the shared keys and values. This allows the full-resolution tokens to be updated by a compact set of descriptors representing the global context.
\begin{equation}
    \mathbf{H} = \text{CrossAttn}(\mathbf{Q}=\mathbf{G}, \mathbf{KV}=\mathbf{D}).
\end{equation}
Crucially, the operation maintains a \textit{global receptive field}, preserving the model's ability to capture long-range dependencies across all input images.

\noindent \textbf{Complexity.} 
Our design significantly reduces the computational complexity of the global block. 
The standard self-attention in Eq.~(2) requires $\mathcal{O}(K^2) = \mathcal{O}(S^2 N^2)$ operations. Our descriptor-based cross-attention reduces this to:
$\mathcal{O}(K \times K_d) = \mathcal{O}(S^2 N^2 / r^2)$, 
With $r=4$ as in our experiments, the complexity reduction is about 16 $\times$.

\subsection{Chunk-Recursive Inference}
\label{ssec:chunk_recursive}

To scale reconstruction to sequences that exceed GPU memory constraints, we propose a chunk-recursive inference scheme. This method processes long sequences sequentially while maintaining a global context across all previously seen chunks, as shown in Fig.~\ref{fig:figure4}.

\input{figures/figure4/figure4}

\noindent \textbf{Problem Formulation.}
Let the input sequence of $S$ images be divided into $T$ consecutive chunks, $\{\mathcal{C}_1, \mathcal{C}_2, \dots, \mathcal{C}_T\}$. For the $t$-th chunk ($1 \leq t \leq T$), let $\mathbf{D}_t$ denote the descriptor tokens generated from chunk $\mathcal{C}_t$ using the method described in Section~\ref{ssec:efficient_attn}. We maintain a set of memory tokens $\mathbf{M}_t$ that accumulates global information from all chunks processed up to step $t$, initialized as $\mathbf{M}_0 = \emptyset$.

\input{tables/table1+2}

\noindent \textbf{Descriptor Attention with Memory.}
For chunk $t$, the global attention computation incorporates historical context through a memory mechanism that maintains information from all previously processed chunks. 
The queries remain the full-resolution image tokens $\mathbf{G}_t$ from the current chunk $\mathcal{C}_t$. The keys and values are formed by concatenating the current chunk's descriptors $\mathbf{D}_t$ with the memory tokens from previous chunks $\mathbf{M}_{t-1}$:
\begin{equation}
\mathbf{H}_t = \text{CrossAttn}\left(\mathbf{Q}=\mathbf{G}_t,\ \mathbf{KV}=[\mathbf{M}_{t-1}, \mathbf{D}_t]\right), 
\end{equation}
where $[\cdot, \cdot]$ denotes the concatenation operation along the sequence dimension.
This design enables each token in the current chunk to attend to both the locally compressed context ($\mathbf{D}_t$) and the globally accumulated history ($\mathbf{M}_{t-1}$), effectively maintaining a global receptive field across the entire sequence while operating on individual chunks. 

\noindent \textbf{Memory Update.}
After processing chunk $t$, the memory is updated by appending the current chunk's descriptors. To constrain memory growth during long sequences, we implement a dropping mechanism that retains only the descriptor tokens from every $p$-th frame. Let $\mathbf{D}_t^{\text{retain}} = \mathbf{D}_t[::p]$ denote the subset of descriptors from every $p$-th frame within the current chunk. The memory is then updated as:
\begin{equation}
\mathbf{M}_t = [\mathbf{M}_{t-1}, \mathbf{D}_t^{\text{retain}}]
\end{equation}
This selective update rule ensures the memory $\mathbf{M}_t$ compactly represents the entire sequence history while limiting its size to grow sublinearly with the number of frames.

\noindent \textbf{Complexity.}
Our chunk-recursive scheme achieves substantial memory efficiency gains over the naive KV-caching in StreamVGGT~\cite{zhuo2025streamvggt}. While StreamVGGT's memory usage is $\mathcal{O}(KL)$, where $L$ is the number of global attention blocks, our approach reduces this to $\mathcal{O}(KL/(pr^2))$ through descriptor compression ($r$) and memory dropping ($p$). 

\subsection{Model Training}

\noindent \textbf{Training Strategy.}
Our training follows a two-stage curriculum. The first stage trains the model on 2-24 randomly shuffled views following VGGT's procedure. The second stage fine-tunes the model on ordered sequences to enable chunk-recursive inference, applying a causal mask to the global attention block that restricts each image to attend only to previous frames in the sequence. 
Unless otherwise specified, we use a spatial compression ratio of $r=4$, a memory drop ratio of $p=5$, and select a key frame every 200 images across all experiments.



\noindent \textbf{Training Data.}
We train our model on seven datasets, which is a subset of VGGT's training data, covering diverse scenarios including synthetic/real-world data, scene-level/object-centric configurations, and indoor/outdoor environments. Specifically, we use BlendedMVS~\cite{yao2020blendedmvs}, CO3Dv2~\cite{reizenstein2021co3d}, ScanNet~\cite{dai2017scannet}, Mapillary~\cite{neuhold2017mapillary}, Arkitscenes~\cite{baruch2021arkitscenes}, MVSSynth~\cite{MVSSynth}, and VirtualKitti~\cite{cabon2020virtualkitti}.

%% file: figures/figure4/figure4.tex
\begin{figure}[!t]
    \centering
    \includegraphics[width=0.99\linewidth]{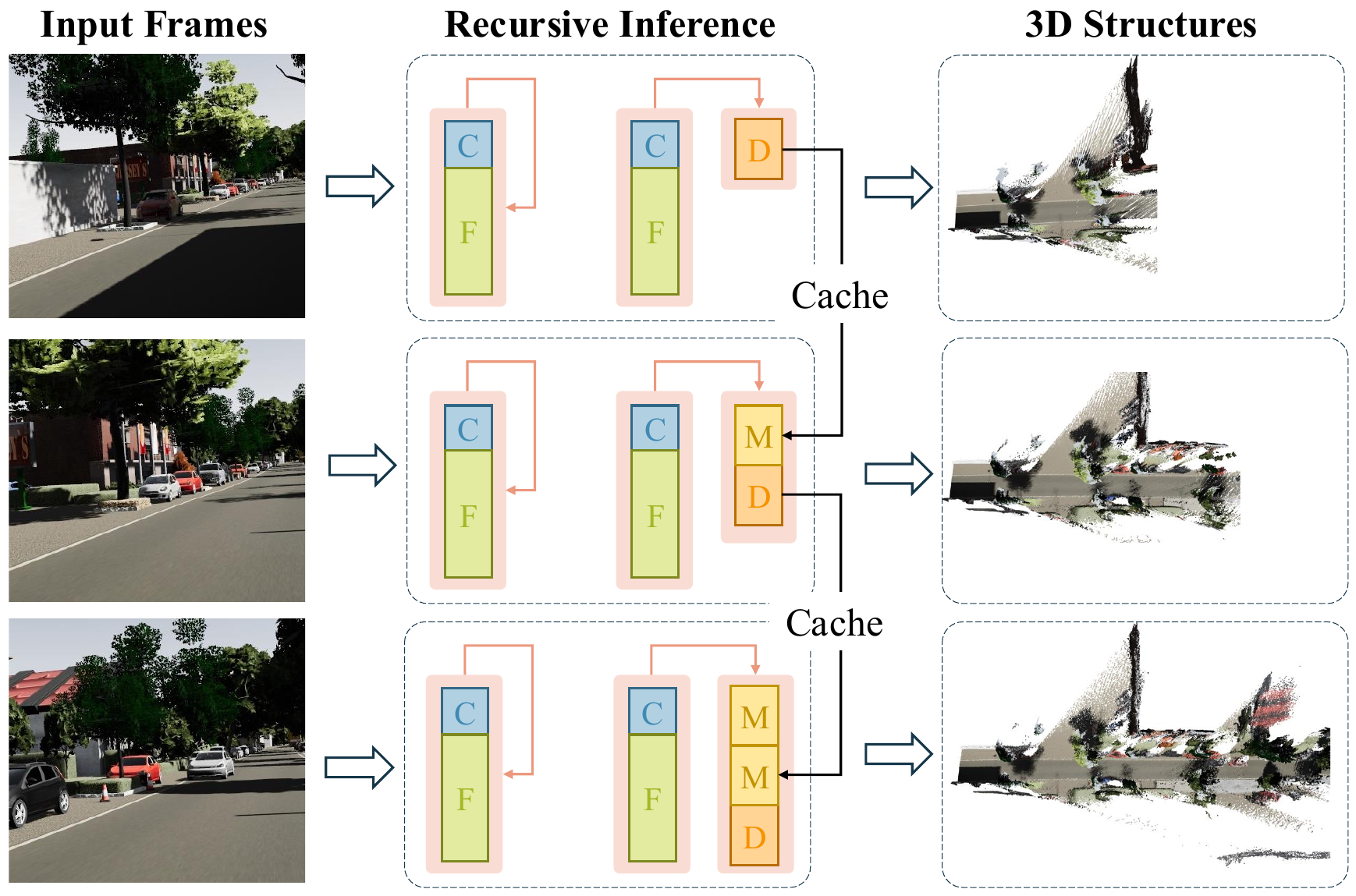}
    \caption{
    \textbf{Chunk Recursive Inference.}
    For long input sequences, we process them in a chunked manner while retaining global reception by caching descriptor tokens. 
    }
    \label{fig:figure4}
    \vspace{-5pt}
\end{figure}

%% file: tables/table1+2.tex
\begin{table*}[!t]
\centering
\begin{minipage}[t]{0.49\textwidth}
\centering
\caption{\textbf{Camera Pose Estimation on RealEstate10K~\cite{zhou2018realestate} and Co3Dv2~\cite{reizenstein2021co3d}.}}
\resizebox{1\textwidth}!{\begin{tabular}{l|ccc|ccc}
\toprule[1.5pt]
\multirow{2}{*}{\textbf{Method}} &
\multicolumn{3}{c}{\textbf{RealEstate10K (unseen)}} &
\multicolumn{3}{c}{\textbf{Co3Dv2}} \\
\cmidrule(r){2-4} \cmidrule(r){5-7}
& RRA@30$\uparrow$ & RTA@30$\uparrow$ & AUC@30$\uparrow$ & RRA@30$\uparrow$ & RTA@30$\uparrow$ & AUC@30$\uparrow$ \\
\midrule
Fast3R~\cite{yang2025fast3r} & 99.05 & 81.86 & 61.68 & 97.49 & 91.11 & 73.43\\
CUT3R~\cite{wang2025cut3r} & 99.82 & 95.10 & 81.47 & 96.19 & 92.69 & 75.82 \\
FLARE~\cite{zhang2025flare} & 99.69 & 95.23 & 80.01 & 96.38 & 93.76 & 73.99 \\
VGGT~\cite{wang2025vggt} & \textbf{99.97} & \textbf{96.22} & \textbf{85.32} & \textbf{98.96} & \textbf{97.13} & \textbf{88.59} \\
FastVGGT~\cite{shen2025fastvggt} & \underline{99.92} & 94.76 & 84.37 & 97.51 & 96.01 & 86.55 \\
\midrule
\textbf{FlashVGGT} & 99.92 & \underline{95.61} & \underline{85.30} & \underline{98.23} & \underline{96.75} & \underline{86.88} \\
\bottomrule[1.5pt]
\end{tabular}}
\label{tab:camera_results}
\end{minipage}
\hfill
\begin{minipage}[t]{0.49\textwidth}
\centering
\caption{\textbf{Monocular Depth Estimation on Sintel~\cite{Butler2012sintel}, Bonn~\cite{palazzolo2019bonn} and NYU-v2~\cite{Silberman2012nyuv2}.} }
\resizebox{0.96\textwidth}!{\begin{tabular}{l|cc|cc|cc}
\toprule[1.5pt]
\multirow{2}{*}{\textbf{Method}} &
\multicolumn{2}{c}{\textbf{Sintel}} &
\multicolumn{2}{c}{\textbf{Bonn}} &
\multicolumn{2}{c}{\textbf{NYU-v2}} \\
\cmidrule(r){2-3} \cmidrule(r){4-5} \cmidrule(r){6-7}
& Abs Rel$\downarrow$ & $\tau<1.25\uparrow$ & Abs Rel$\downarrow$ & $\tau<1.25\uparrow$ & Abs Rel$\downarrow$ & $\tau<1.25\uparrow$ \\
\midrule
Fast3R~\cite{yang2025fast3r} & 0.544 & 0.509 & 0.169 & 0.796 & 0.093 & 0.898 \\
CUT3R~\cite{wang2025cut3r} & 0.418 & 0.520 & 0.058 & 0.967 & 0.081 & 0.914 \\
FLARE~\cite{zhang2025flare} & 0.606 & 0.402 & 0.130 & 0.836 & 0.089 & 0.898 \\
VGGT~\cite{wang2025vggt} & \textbf{0.335} & \textbf{0.599} & \textbf{0.053} & \textbf{0.970} & \textbf{0.056} & \textbf{0.951} \\
FastVGGT~\cite{shen2025fastvggt} & \underline{0.337} & 0.582 & 0.056 & 0.952 & 0.058 & 0.943 \\
\midrule
\textbf{FlashVGGT} & 0.346 & \underline{0.586} & \underline{0.054} & \underline{0.957} & \underline{0.058} & \underline{0.947} \\
\bottomrule[1.5pt]
\end{tabular}}
\label{tab:depth_results}
\end{minipage}
\end{table*}

%% file: sec/4-experiment.tex
\section{Experiment}
\input{tables/table3}

We evaluate FlashVGGT against state-of-the-art methods across three core tasks: (i) monocular and sparse-view reconstruction (Sec.~\ref{mono_sparse}), long-sequence dense 3D reconstruction (Sec.~\ref{dense_recon}), and (iii) online dense 3D reconstruction (Sec.~\ref{online_recon}). We subsequently analyze the effectiveness of our key design choices in Sec.~\ref{analysis}. All evaluations are conducted on a single NVIDIA H800 GPU.

\subsection{Monocular and Sparse Reconstruction} \label{mono_sparse}
\noindent \textbf{Camera Pose Estimation.}
Following \cite{wang2025vggt}, we evaluate our method on the CO3Dv2~\cite{reizenstein2021co3d} and RealEstate10K~\cite{zhou2018realestate} datasets for camera pose estimation on short sequences. 
For each scene, we randomly select 10 images and report standard metrics: RRA (Relative Rotation Accuracy) and RTA (Relative Translation Accuracy), and AUC, the area under the accuracy-threshold curve for the minimum of RRA and RTA.
As shown in Tab.~\ref{tab:camera_results}, FlashVGGT achieves highly competitive performance.
On the out-of-distribution RealEstate10K dataset, our method closely matching VGGT's metrics, while being better than the concurrent work FastVGGT~\cite{shen2025fastvggt}. 
On the CO3Dv2 dataset, FlashVGGT also significantly outperforms all other efficient methods like FastVGGT~\cite{shen2025fastvggt} and CUT3R~\cite{wang2025cut3r}, and remains within a narrow margin of the original VGGT. 

\noindent \textbf{Monocular Depth Estimation.}
We evaluate single-image depth prediction on the Sintel~\cite{Butler2012sintel}, Bonn~\cite{palazzolo2019bonn}, and NYU-v2~\cite{Silberman2012nyuv2} datasets, reporting standard metrics: Absolute Relative Error (Abs Rel) and accuracy under threshold $\tau<1.25$. As shown in Tab.~\ref{tab:depth_results}, FlashVGGT demonstrates strong performance across all benchmarks. 
On Sintel, our method achieves competitive results that are comparable with VGGT and significantly outperform other efficient methods. 
For the Bonn and NYU-v2 datasets, FlashVGGT consistently ranks as the second-best method, closely following VGGT within a small gap while substantially outperforming FastVGGT and other concurrent approaches. 

These results demonstrate that our efficient descriptor-based attention preserves the strong geometric reasoning capabilities of VGGT, while our subsequent experiments will show it does so at a fraction of the computational cost.

\subsection{Long-sequence Dense 3D Reconstruction}\label{dense_recon}
This section presents our main results on long-sequence dense 3D reconstruction. 
We conduct a thorough evaluation of state-of-the-art offline methods capable of processing long sequences, comparing against Fast3R~\cite{yang2025fast3r}, VGGT~\cite{wang2025vggt}, and FastVGGT~\cite{shen2025fastvggt}. For VGGT, we adopt the memory-efficient implementation from~\cite{shen2025fastvggt}, which maintains the original accuracy while enabling inference on sequences of over 1,000 images.
Following \cite{shen2025fastvggt}, we benchmark performance across totally 107 scenes from N-RGBD~\cite{azinovic2022nrgbd}, 7-Scenes~\cite{shotton2013sevenscene}, and a subset of ScanNet~\cite{dai2017scannet} on 100, 500 and 1,000 images respectively.
We evaluate each method comprehensively across 12 metrics covering depth estimation, point cloud reconstruction and camera pose estimation. 
For depth estimation, we report Absolute Relative Error (Abs Rel) and $\tau<1.25$ ratio.
For point cloud reconstruction, we report Accuracy (Acc), Completeness (Comp), Chamfer Distance (CD), and Normal Consistency (NC).
For camera estimation, we report Absolute Translation Error (APE), Absolute Rotation Error (ARE), Relative Translation Error (RPE-Trans), and Relative Rotation Error (RPE-Rot).
Additionally, we report inference time and maximum GPU memory usage to provide a complete picture of each method's practical utility.

The comprehensive results in Tab.~\ref{tab:dense-recon} demonstrate FlashVGGT's superior scalability and efficiency while maintaining competitive reconstruction quality across varying sequence lengths. For 100-image sequences, FlashVGGT achieves performance comparable to VGGT while being over 3$\times$ faster. At 500 images, it remains highly competitive with VGGT while achieving an over 8$\times$ speedup. For 1,000-image sequences, VGGT suffers from noticeable performance degradation due to attention dilution across excessive tokens, whereas FlashVGGT maintains high accuracy with over 10$\times$ faster inference. Across all sequence lengths, FlashVGGT consistently outperforms other efficient methods like FastVGGT~\cite{shen2025fastvggt} and Fast3R~\cite{yang2025fast3r}, establishing a superior balance between efficiency and reconstruction fidelity.
Fig.~\ref{fig:figure5} presents a qualitative comparison of FlashVGGT against other methods. The results demonstrate that FlashVGGT produces more complete and robust reconstructions from long input sequences while achieving significantly faster inference. 
Notably, VGGT exhibits substantial performance degradation as sequence length increases, as seen in the room reconstruction from 1,000 images.
We attribute this failure to noisy and redundant interactions over extremely long input (over 1M tokens for 1,000 images). 
In contrast, FlashVGGT avoids this pitfall by learning a compact, stable set of descriptor tokens that distill essential information, maintaining consistent performance across long sequences.

\input{figures/figure5/figure5}

\subsection{Online Dense 3D Reconstruction}\label{online_recon}

We evaluate FlashVGGT in an online inference setting with a chunk size of 10 images against three recent online reconstruction methods: CUT3R~\cite{wang2025cut3r}, TTT3R~\cite{chen2025ttt3r}, and StreamVGGT~\cite{zhuo2025streamvggt}. Experiments are conducted on N-RGBD~\cite{azinovic2022nrgbd} using 500-image sequences from each scene.
As shown in Tab.~\ref{tab:online}, our method significantly outperforms previous approaches across all metrics. FlashVGGT achieves the best reconstruction quality while being over 3.3$\times$ faster than the fastest competitor CUT3R. Notably, we achieve this while using less than a quarter of the memory required by StreamVGGT.
Qualitative results in Fig.~\ref{fig:figure6} further demonstrate our approach's superiority. FlashVGGT successfully reconstructs complete room geometry and fine details (e.g., tiny objects on the table), while CUT3R and TTT3R suffer from accumulated errors and fail to recover meaningful structures. Although StreamVGGT produces reasonable geometry, it requires over 20$\times$ more time and still fails to capture fine details effectively.
\input{tables/table4}
\input{figures/figure6/figure6}

\subsection{Model Analysis and Discussion}\label{analysis}

\noindent \textbf{Spatial Compression Methods.} \label{compress_method}
We evaluated five strategies for producing descriptor tokens: average pooling, top-k selection based on token norm, nearest-neighbor interpolation, bilinear interpolation, and a lightweight learnable compressor consisting of a depth-wise convolution followed by a point-wise linear layer (Table \ref{tab:spatial_compression}).
Our analysis reveals that interpolation-based methods consistently outperform other approaches.
We argue that their edge stems from locality preservation: as the DINO encoder outputs tokens corresponding to 14$\times$14 pixel patches, aggressive aggregation methods such as pooling merge information from many distant patches and wash out fine-grained cues. Interpolation, instead, blends only a handful of spatially adjacent tokens with distance-aware weights, retaining high-frequency detail that downstream tasks find useful.
While top-k selection preserves original token values, it relies on the assumption that larger token norms correlate with informative local descriptors, which may not always hold.
Finally, the learnable compressor neither improves quality nor stability; its limited capacity appears insufficient to capture the rich spatial patterns of the descriptors.

\input{tables/table5}

\noindent \textbf{Spatial Compression Ratio.}
Fig.~\ref{fig:figure9} illustrates the trade-off between reconstruction accuracy (Chamfer Distance, left axis) and inference speed (right axis) across different compression ratios 
$r$. While larger $r$ values yield faster inference, they also lead to a progressive decline in accuracy as fine-grained spatial information is lost. The ratio $r=4$ provides an optimal balance, offering a significant speedup with minimal loss in reconstruction quality. Beyond this point, the rate of performance degradation increases substantially for diminishing gains in speed.
\input{figures/figure9/figure9}

\noindent \textbf{Auxiliary Descriptor Tokens.}
We identify that including the auxiliary descriptor tokens as described in Sec.~\ref{aux_tokens} is crucial for maintaining reconstruction quality, as they augment the fine-grained information losses in compression.
As shown in Fig.~\ref{fig:figure8}, omitting these auxiliary tokens degrades global geometric consistency. This effect is particularly pronounced in long sequences with low inter-frame overlap, \eg, autonomous driving scenarios.
\input{figures/figure8/figure8}

\input{figures/figure7/figure7}
\noindent \textbf{Confidence Maps.} \label{Confidence_pattern}
Fig.~\ref{fig:figure7} shows a comparison of the confidence maps predicted by VGGT and our method.
VGGT tends to produce over-confident predictions, assigning disproportionately low confidence scores to homogeneous regions such as walls and computer screens. This often results in gaps and holes in the final reconstruction after filtering out low-confidence points.
In contrast, our method generates a more calibrated and spatially coherent confidence map. This allows for the effective preservation of structural details while robustly filtering out noise, leading to more complete and reliable 3D reconstructions.


%% file: tables/table3.tex
\begin{table*}[]
\centering
\caption{\textbf{Large-Scale Dense 3D Reconstruction.} 
Evaluation across 100, 500, and 1,000 image sequences, with results averaged over N-RGBD~\cite{azinovic2022nrgbd}, 7-Scenes~\cite{shotton2013sevenscene}, and ScanNet~\cite{dai2017scannet}. Point cloud and camera pose metrics are multiplied by 100 for better readability. }
\label{tab:dense-recon}
\resizebox{\textwidth}{!}{%
\begin{tabular}{@{}c|l|cc|cccc|cccc|cc@{}}
\toprule[1.5pt]
\multirow{2}{*}{Frames} &
  \multirow{2}{*}{Method} &
  \multicolumn{2}{c|}{\textbf{Depth}} &
  \multicolumn{4}{c|}{\textbf{Point}} &
  \multicolumn{4}{c|}{\textbf{Camera}} &
  \multicolumn{2}{c}{\textbf{Resource}} \\ \cmidrule(l){3-14} 
 &
   &
  Abs Rel$\downarrow$ &
  $\tau < 1.25$$\uparrow$ &
  Acc$\downarrow$ &
  Comp$\downarrow$ &
  CD$\downarrow$ &
  NC$\uparrow$ &
  APE$\downarrow$ &
  ARE$\downarrow$ &
  RPE-Trans$\downarrow$ &
  RPE-Rot$\downarrow$ &
  Time (s)$\downarrow$ &
  Mem. (GB)$\downarrow$ \\ \midrule
\multirow{4}{*}{100} &
  Fast3R~\cite{yang2025fast3r} &
  0.038 &
  0.951 &
  1.164 &
  1.900 &
  1.532 &
  62.10 &
  2.654 &
  3.123 &
  0.494 &
  0.756 &
  4.40 &
  13.94 \\
 &
  VGGT~\cite{wang2025vggt} &
  {\ul 0.029} &
  0.983 &
  {\ul 0.962} &
  1.162 &
  1.062 &
  \textbf{72.48} &
  \textbf{1.537} &
  {\ul 2.935} &
  \textbf{0.353} &
  \textbf{0.493} &
  4.93 &
  {\ul 12.26} \\
 &
  FastVGGT~\cite{shen2025fastvggt} &
  0.029 &
  {\ul 0.984} &
  0.988 &
  \textbf{1.092} &
  \textbf{1.040} &
  68.34 &
  1.663 &
  3.011 &
  0.507 &
  0.702 &
  {\ul 2.74} &
  12.68 \\ \cmidrule(l){2-14} 
 &
  \textbf{FlashVGGT} &
  \textbf{0.028} &
  \textbf{0.990} &
  \textbf{0.897} &
  {\ul 1.142} &
  {\ul 1.019} &
  {\ul 70.14} &
  {\ul 1.648} &
  \textbf{2.834} &
  {\ul 0.447} &
  {\ul 0.621} &
  \textbf{1.54} &
  \textbf{12.07} \\ \midrule
\multirow{4}{*}{500} &
  Fast3R~\cite{yang2025fast3r} &
  0.045 &
  0.962 &
  1.432 &
  1.590 &
  1.511 &
  58.8 &
  6.784 &
  8.570 &
  2.343 &
  2.120 &
  62.40 &
  33.30 \\
 &
  VGGT~\cite{wang2025vggt} &
  0.035 &
  0.967 &
  1.484 &
  \textbf{1.209} &
  1.347 &
  \textbf{71.15} &
  {\ul 4.414} &
  \textbf{6.855} &
  \textbf{1.453} &
  \textbf{1.558} &
  90.97 &
  37.22 \\
 &
  FastVGGT~\cite{shen2025fastvggt} &
  {\ul 0.034} &
  {\ul 0.967} &
  {\ul 1.388} &
  {\ul 1.241} &
  {\ul 1.314} &
  66.70 &
  4.561 &
  7.064 &
  1.722 &
  1.952 &
  {\ul 29.04} &
  {\ul 39.33} \\ \cmidrule(l){2-14} 
 &
  \textbf{FlashVGGT} &
  \textbf{0.034} &
  \textbf{0.969} &
  \textbf{1.314} &
  1.283 &
  \textbf{1.298} &
  {\ul 70.18} &
  \textbf{4.298} &
  {\ul 6.950} &
  {\ul 1.474} &
  {\ul 1.576} &
  \textbf{12.54} &
  \textbf{33.39} \\ \midrule
\multirow{4}{*}{1000} &
  Fast3R~\cite{yang2025fast3r} &
  0.122 &
  0.855 &
  3.076 &
  1.457 &
  2.267 &
  52.5 &
  12.67 &
  22.36 &
  9.530 &
  11.34 &
  224.10 &
  61.95 \\
 &
  VGGT~\cite{wang2025vggt} &
  0.048 &
  0.951 &
  2.039 &
  1.004 &
  1.521 &
  {\ul 68.65} &
  6.519 &
  15.80 &
  {\ul 2.222} &
  7.029 &
  372.80 &
  68.40 \\
 &
  FastVGGT~\cite{shen2025fastvggt} &
  {\ul 0.034} &
  {\ul 0.986} &
  {\ul 1.322} &
  {\ul 1.089} &
  {\ul 1.206} &
  66.05 &
  {\ul 5.651} &
  {\ul 8.400} &
  2.553 &
  {\ul 2.898} &
  {\ul 78.22} &
  {\ul 72.60} \\ \cmidrule(l){2-14} 
 &
  \textbf{FlashVGGT} &
  \textbf{0.032} &
  \textbf{0.991} &
  \textbf{1.160} &
  \textbf{1.096} &
  \textbf{1.128} &
  \textbf{69.63} &
  \textbf{5.237} &
  \textbf{8.242} &
  \textbf{2.067} &
  \textbf{2.802} &
  \textbf{35.32} &
  \textbf{60.74} \\ \bottomrule[1.5pt]
\end{tabular}%
}
\end{table*}

%% file: figures/figure5/figure5.tex
\begin{figure*}[!t]
    \centering
    \includegraphics[width=0.95\linewidth]{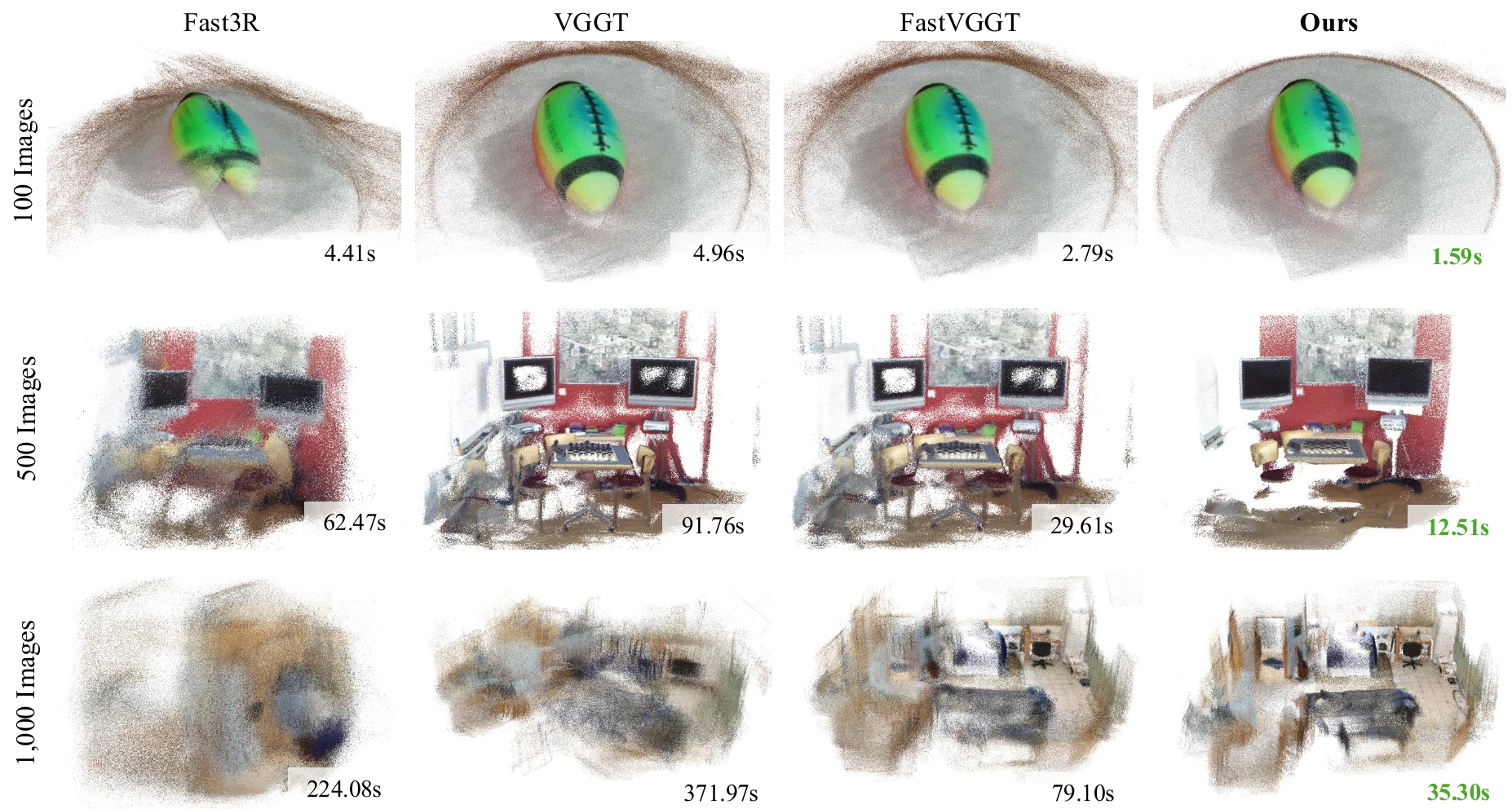}
    \caption{
    \textbf{Qualitative comparison of long-sequence dense 3D reconstruction.} 
    FlashVGGT produces more robust reconstructions over long input sequences while being significantly faster.
    Only points with top 90\% confidence are shown for better visualization. 
    }
    \vspace{-5pt}
    \label{fig:figure5}
\end{figure*}

%% file: tables/table4.tex
\begin{table}[]
\centering
\caption{\textbf{Online Dense 3D Reconstruction on N-RGBD~\cite{azinovic2022nrgbd}.}}
\label{tab:online}
\resizebox{\columnwidth}{!}{%
\begin{tabular}{@{}l|cccccc@{}}
\toprule[1.5pt]
Method                               & Abs Rel$\downarrow$ & Acc$\downarrow$ & Comp$\downarrow$ & APE$\downarrow$ & Time (s)       & Mem (GB)    \\ \midrule
CUR3R~\cite{wang2025cut3r} & 0.375 & 4.890 & 3.426 & 23.456 & {\ul 34.19} & \textbf{6.16} \\
TTT3R~\cite{chen2025ttt3r} & 0.134 & 3.567 & 1.954 & 16.434 & 35.67       & \textbf{6.16} \\
StreamVGGT~\cite{zhuo2025streamvggt} & {\ul 0.086}         & {\ul 2.456}     & {\ul 1.235}      & {\ul 6.543}     & 209.50         & 70.70       \\ \midrule
\textbf{FlashVGGT}                   & \textbf{0.047}      & \textbf{1.912}  & \textbf{0.625}   & \textbf{4.792}  & \textbf{12.52} & {\ul 13.10} \\ \bottomrule[1.5pt]
\end{tabular}%
}
\vspace{-5pt}
\end{table}

%% file: figures/figure6/figure6.tex
\begin{figure}[htp]
    \centering
    \includegraphics[width=0.95\linewidth]{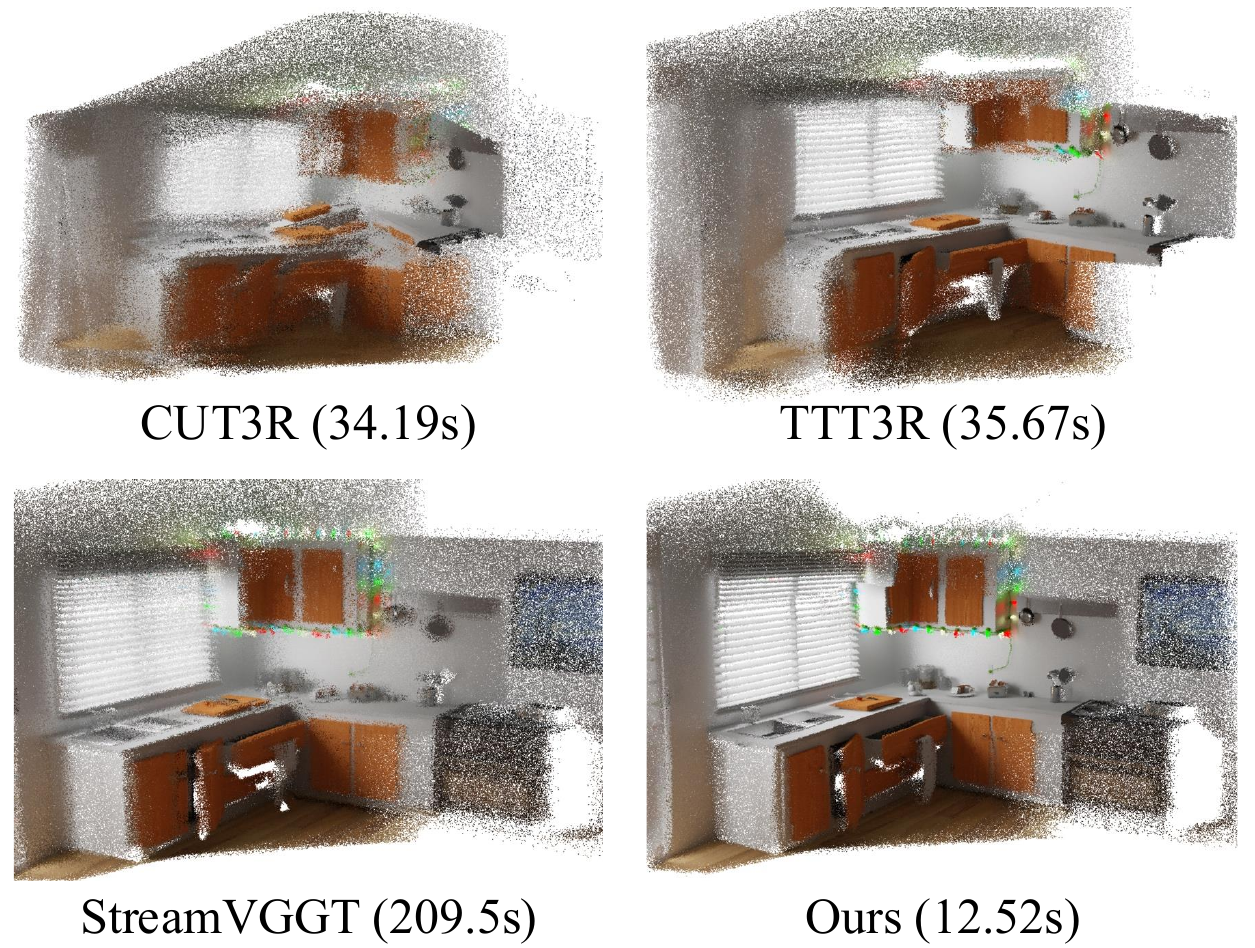}
    \caption{
    \textbf{Qualitative comparison of online 3D reconstruction.} 
    All methods are evaluated on a sequence of 500 images.
    }
    \label{fig:figure6}
    \vspace{-8pt}
\end{figure}

%% file: tables/table5.tex
\begin{table}[]
\centering
\caption{\textbf{Comparison of different spatial compression techniques.} Evaluated on N-RGBD~\cite{azinovic2022nrgbd} with 100 input images.}
\label{tab:spatial_compression}
\resizebox{0.9\columnwidth}{!}{%
\begin{tabular}{@{}l|cccccc@{}}
\toprule
         & Abs Rel$\downarrow$ & Acc$\downarrow$ & Comp$\downarrow$ & NC$\uparrow$   & APE$\downarrow$ & ARE$\downarrow$ \\ \midrule
Pooling  & 0.019               & 0.560           & 0.301            & 75.68          & 2.256           & 4.008           \\
Top-k    & 0.019               & 0.569           & 0.331            & 75.13          & 2.234           & 4.516           \\
Learned  & 0.023               & 0.643           & 0.675            & 68.33          & 2.658           & 5.183           \\
Nearest  & 0.014      & 0.441           & 0.273            & 76.96          & 1.902           & 3.456           \\ \midrule
Bilinear & \textbf{0.014}      & \textbf{0.436}  & \textbf{0.272}   & \textbf{77.75} & \textbf{1.890}  & \textbf{3.438}  \\ \bottomrule
\end{tabular}%
}
\end{table}

%% file: figures/figure9/figure9.tex
\begin{figure}[thp]
    \centering
    \includegraphics[width=0.95\linewidth]{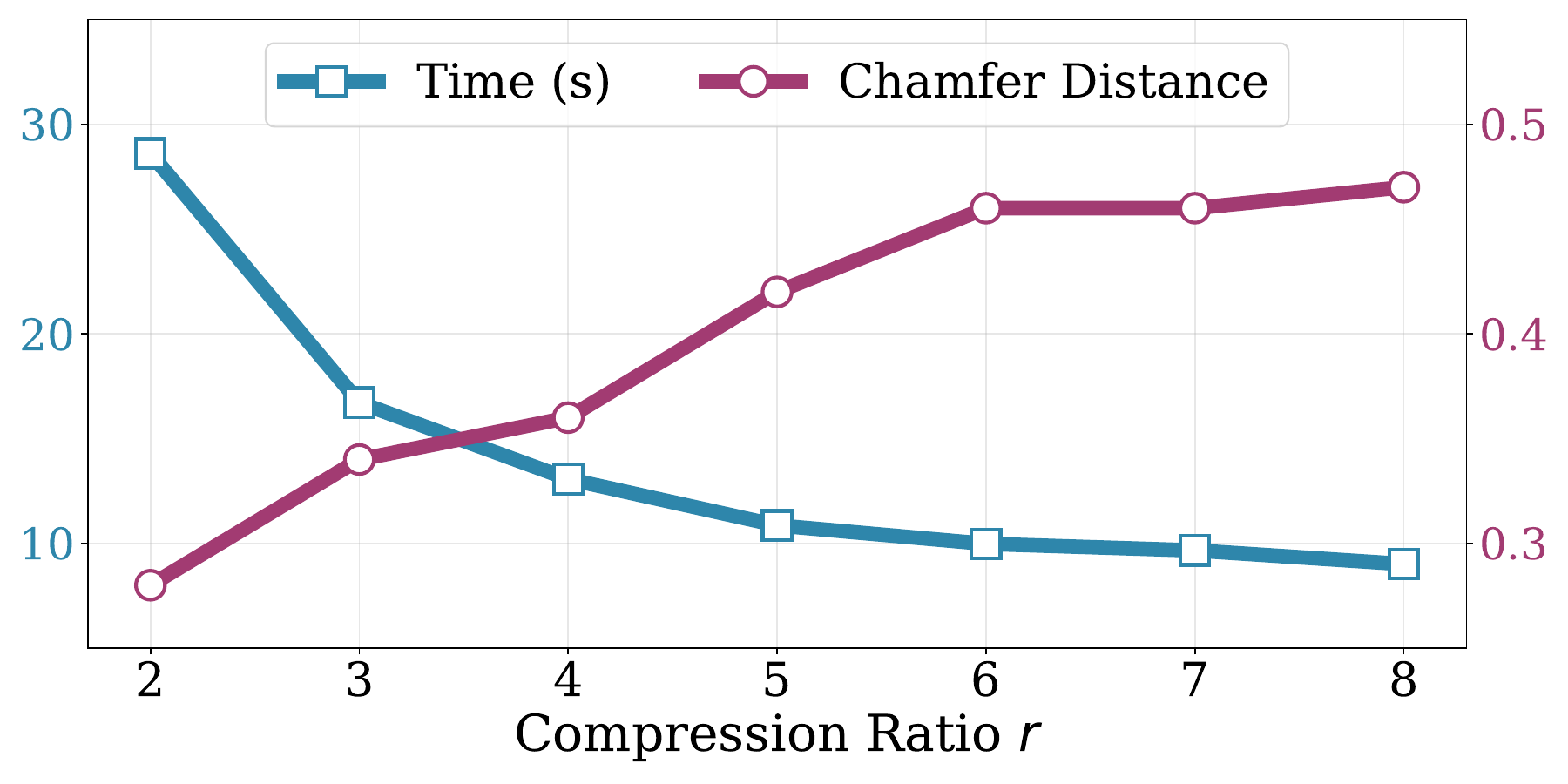}
    \caption{
    \textbf{The impact of different compression ratios.}
    The ratio of 4 provides a balanced choice between quality and speed.
    }
    \label{fig:figure9}
\end{figure}

%% file: figures/figure8/figure8.tex
\begin{figure}[thp]
    \centering
    \includegraphics[width=0.99\linewidth]{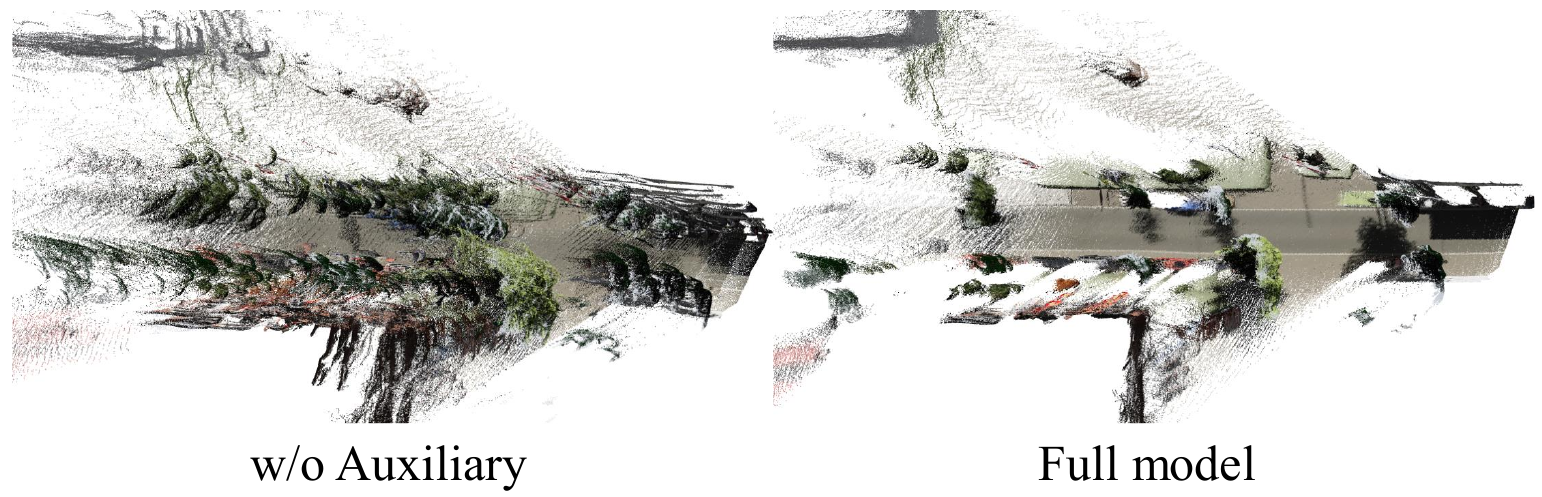}
    \caption{
    \textbf{The impact of auxiliary descriptor tokens.} Including auxiliary tokens improves geometric quality.
    }
    \label{fig:figure8}
\end{figure}

%% file: figures/figure7/figure7.tex
\begin{figure}[thp]
    \centering    \includegraphics[width=0.95\linewidth]{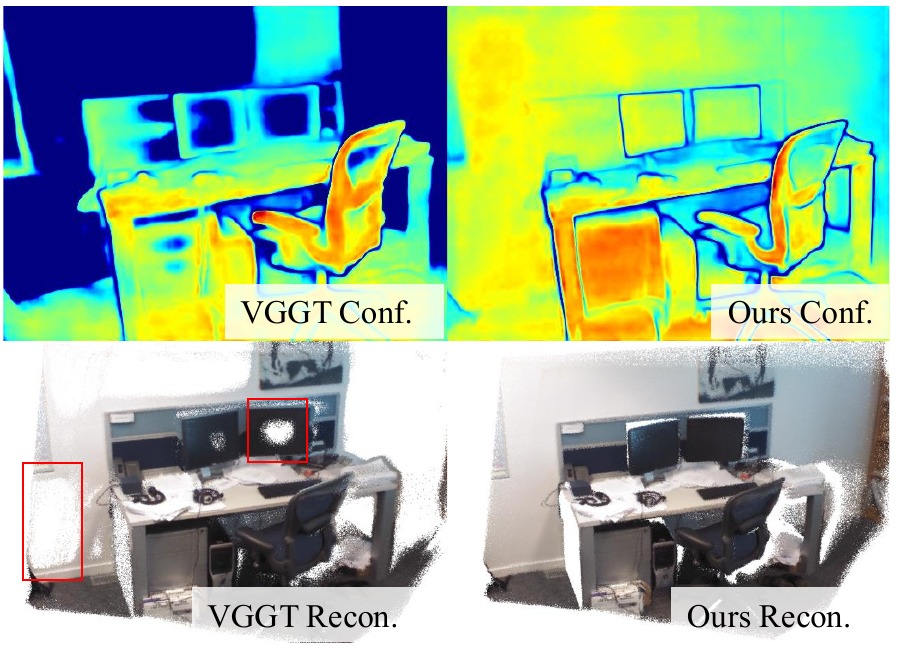}
    \caption{
    \textbf{Analysis of confidence maps.} Ours method produces spatially coherent confidence scores that better preserve planar structures (\eg, the computer screen) while filtering noise.
    }
    \label{fig:figure7}
    \vspace{-1em}
\end{figure}

%% file: sec/5-conclusion.tex
\section{Conclusion}
In this work, we introduced FlashVGGT, a novel framework that overcomes the scalability bottleneck of global attention in feed-forward 3D reconstruction. 
We identified that the full global self-attention in models like VGGT is computationally prohibitive and largely unnecessary.
Our solution centers on a compressed descriptor attention paradigm that replaces dense attention with efficient cross-attention from all image tokens to a compact set of learned descriptor tokens. Furthermore, the compact nature of these descriptors enables a chunk-recursive inference scheme, allowing FlashVGGT to process very long sequence with a manageable memory footprint.
Through extensive experiments, we demonstrated that FlashVGGT achieves a superior balance between efficiency and accuracy. It matches the reconstruction quality of VGGT while achieving over 90\% inference speedup for 1,000-image sequences and scaling effectively to over 3,000 images. By making high-fidelity, large-scale 3D reconstruction both fast and practical, FlashVGGT opens the door for more demanding real-world applications.

\noindent \textbf{Limitations:}
 While highly efficient for long sequences, our method exhibits a slight performance degradation on shorter sequences. Furthermore, the design space for descriptor attention remains largely unexplored. Please refer to the supplementary material for a detailed discussion.

%% file: sec/6-acknowledgment.tex
\section*{Acknowledgments}
The research is supported in part by Early Career Scheme of the Research Grants Council (RGC) of the Hong Kong SAR under grant No. 26202321, ITF PRP/046/24FX, Science \& Technology Cooperation Program of Shandong under grant No. SDST26EG01, SAIL Research Project, HKUST-Zeekr Coolaborative Research Fund, Westwell Project, and Tencent Rhino-Bird Focused Research Program. 

%% file: sec/X_suppl.tex
\clearpage
\setcounter{page}{1}
\maketitlesupplementary
\appendix

\noindent \textbf{Overview.}
The supplementary material is organized as follows: Sec.~\ref{suppl:impl} elaborates on training specifics and hyperparameters. Sec.~\ref{suppl:analysis} presents extended experimental analysis and ablation studies. Finally, Sec.~\ref{suppl:discussion} discusses the limitations of our current work and promising directions for future research.

\section{Implementation Details}\label{suppl:impl}

\noindent \textbf{Model and Optimization.}
We initialize FlashVGGT from a pre-trained VGGT~\cite{wang2025vggt} checkpoint. During training, we freeze the image encoder and reconstruction heads, optimizing only the alternating attention aggregator which comprises approximately 50\% of the total parameters while employing the original VGGT loss functions. We use the Adam-W optimizer~\cite{loshchilov2017adamw} with an initial learning rate of $4\times10^{-6}$, linear warmup, and cosine decay. Both training stages run for 10,000 iterations on 4 H800 GPUs, completing in approximately 16 hours.

\noindent \textbf{Training Protocol.}
We adopt VGGT's dynamic batching scheme, randomly sampling 2 to 24 frames per iteration. Input images are pre-processed by resizing the longer side to 518 pixels while randomizing the aspect ratio between 0.33 and 1.0. We apply standard data augmentation including color jittering and random grayscale conversion. For training stability, we employ gradient norm clipping with a threshold of 1.0, and leverage both bfloat16 precision and gradient checkpointing as in the original VGGT.

\section{Additional Analysis}\label{suppl:analysis}

\noindent \textbf{Ablation of auxiliary descriptor tokens.}
As shown in Tab.~\ref{tab:ablation-aux}, we analyze the contribution of each auxiliary token component using 500-image sequences from 7Scenes~\cite{shotton2013sevenscene}. The full model achieves the best overall performance, validating the importance of all auxiliary tokens.
The most critical component is the reference frame tokens, whose removal causes the most severe degradation, particularly in pose estimation (APE increases by 96\% and ARE by 68\%). This confirms that preserving the full coordinate frame is essential for global geometric consistency.
Camera tokens also prove vital, as their absence leads to noticeable deterioration in reconstruction quality while providing minimal memory savings. This demonstrates that explicit camera parameter representation significantly aids the network's geometric reasoning.
While key frame tokens offer the most modest improvements, they still enhance both reconstruction (CD) and camera pose (APE) metrics with negligible computational overhead. This suggests that distributing full-resolution information across the sequence helps maintain local detail preservation.
Notably, all auxiliary components contribute to performance with minimal impact on efficiency, confirming our design provides an effective accuracy-efficiency trade-off.

\input{tables/suppl/abl_aux}

\noindent \textbf{Key-frame selection methods.}
As shown in Tab.~\ref{tab:abl-keyframe}, we compare different strategies for selecting key frames. The clustering-based approach consistently outperforms both random and fixed-stride selection across all accuracy metrics.
Our proposed clustering method achieves the best performance, demonstrating its ability to select representative frames that better capture the scene's geometric diversity. This comes at a minimal computational cost, adding only 0.29 seconds compared to the fastest method.
While fixed-stride selection offers slightly better efficiency, it shows clear performance limitations, particularly in pose estimation. This suggests that uniformly distributed frames may miss critical viewpoints needed for optimal geometric reconstruction.
Random selection performs similarly to fixed-stride but with more variability across metrics, confirming that naive approaches cannot reliably identify the most informative frames for 3D reconstruction.
The results validate that our clustering-based key frame selection effectively identifies geometrically representative views, providing better reconstruction quality with negligible overhead compared to simpler alternatives.

\input{tables/suppl/abl_keyframe}

\noindent \textbf{Memory retain rate $p$. }
We analyze the trade-off between efficiency and accuracy by varying the memory retain rate $p$, which controls how many historical descriptor tokens are preserved during chunk-recursive inference. As shown in Fig.~\ref{fig:figure10}, lower values of $p$ (more aggressive memory dropping) yield better efficiency but gradually degrade reconstruction quality.
Notably, the performance drop from $p=1$ to $p=5$ is minimal compared to the substantial efficiency gains. This suggests that carefully selected memory dropping can eliminate redundant historical information with negligible impact on reconstruction quality. 
\input{figures/figure10/figure10}

\noindent \textbf{Chunk size. }
We analyze the impact of chunk size on online reconstruction performance in Tab.~\ref{tab:chunk}. The results demonstrate that chunk size has minimal effect on reconstruction quality across all metrics. However, chunk size significantly impacts computational efficiency: increasing from 1 to 100 frames per chunk provides a 2.3$\times$ speedup at the cost of 49\% higher memory usage.
This reveals a flexible trade-off between speed and memory constraints. Applications prioritizing throughput can use larger chunks (\eg, 100) for faster processing, while memory-constrained environments can employ smaller chunks (\eg, 10-50). The stability of reconstruction metrics across chunk sizes confirms the robustness of our chunk-recursive scheme, making it adaptable to diverse deployment scenarios.
\input{tables/suppl/abl_chunksize}

\noindent \textbf{Resources consumption.}
As shown in Tab.~\ref{tab:resources}, FlashVGGT demonstrates substantial efficiency gains across all resource metrics while scaling to longer sequences than competing methods.
\textbf{(i)} Computational Efficiency: FlashVGGT achieves remarkable reductions in both time and FLOPs. For 1,000 images, our method is 10.1$\times$ faster than VGGT~\cite{wang2025vggt} and requires 15.8$\times$ fewer FLOPs. Even compared to FastVGGT~\cite{shen2025fastvggt}, FlashVGGT provides a 2.1$\times$ speedup and 2.1$\times$ FLOP reduction, demonstrating the superiority of our descriptor-based approach over token merging.
\textbf{(ii)} Memory Efficiency: FlashVGGT maintains the lowest memory footprint across all sequence lengths. At 1,000 images, it uses 11\% less memory than VGGT and 16\% less than FastVGGT. This memory advantage enables FlashVGGT to successfully process 1,200-image sequences where both baselines fail due to out-of-memory errors. Note that FastVGGT takes more memory as it needs to compute token similarity across all tokens.
\textbf{(iii)} Scalability: The computational advantages become more pronounced with longer sequences. While VGGT and FastVGGT cannot process beyond 1,000 images, FlashVGGT maintains efficient operation at 1,200 images with only 51.25 seconds and 71.61GB memory, demonstrating robust scalability for large-scale reconstruction tasks.
These results confirm that our descriptor-based attention mechanism provides fundamental improvements in computational efficiency without compromising on reconstruction quality, enabling practical processing of very long image sequences.
\input{tables/suppl/resources}

\noindent \textbf{Comparison with latent cross-attention.}
While our method shares the high-level intuition of reducing computation via asymmetric attention with latent cross-attention methods like Perceiver~\cite{jaegle2021perceiver}, our approach differs fundamentally. Unlike Perceiver-style methods that use randomly initialized learnable tokens as queries to aggregate information from the input, we use the original input tokens as queries and a spatially compressed version as keys and values. This design preserves the original input resolution, making it more suitable for dense prediction tasks like 3D reconstruction. Furthermore, our compressed descriptors carry strong data-dependent priors through spatial resampling, maintaining the input's structural distribution rather than learning a generic latent representation.
To validate our approach, we compare against a Perceiver-style alternative that uses additional learnable latent tokens per frame. During frame attention, these latent tokens interact with tokens in the same frame, while in global attention, we compute cross-attention from frame tokens to these latent tokens. As shown in Tab.~\ref{tab:perceiver}, our method significantly outperforms the Perceiver-style approach across all metrics. These results demonstrate that leveraging data-dependent compression is crucial for high-quality 3D reconstruction.

\input{tables/suppl/perceiver}

\noindent \textbf{In-the-wild Benchmark.}
We evaluated our model on the IMC PhotoTourism benchmark. Although our architecture is primarily optimized for long sequences, it exhibits only a minor accuracy gap compared to VGGT on these shorter in-the-wild sequences (5-25 frames) and consistently outperforms the concurrent efficient alternative, FastVGGT.
\begin{table}[h]
\centering
\caption{Evaluation on the IMC PhotoTourism benchmark.}
\resizebox{0.8\columnwidth}{!}{%
\begin{tabular}{@{}l|cccc@{}}
\toprule
          & AUC@3          & AUC@5          & AUC@10         & Time (s)      \\ \midrule
VGGT      & \textbf{39.23} & \textbf{52.74} & \textbf{71.26} & 0.37          \\
FastVGGT  & 38.58          & 51.43          & 70.12          & {\ul 0.35}    \\ \midrule
FlashVGGT & {\ul 38.62}    & {\ul 51.87}    & {\ul 70.49}    & \textbf{0.26} \\ \bottomrule
\end{tabular}%
}
\end{table}

\noindent \textbf{Stricter Camera Pose Metrics.}
We evaluated our model with stricter thresholds of 5 and 10 degrees on RealEstate10K (10 frames per sequence). Although our architecture is primarily optimized for long sequences, it closely matches VGGT’s metrics on these short sequences, while performing better than the concurrent work, FastVGGT~\cite{shen2025fastvggt}.

\begin{table}[h]
\centering
\caption{Camera pose evaluation with stricter thresholds.}
\resizebox{\columnwidth}{!}{%
\begin{tabular}{@{}l|ccccccl@{}}
\toprule
          & Racc@5         & Tacc@5         & Auc@5          & Racc@10        & Tacc@10        & Auc@10         & Times (s)     \\ \midrule
VGGT      & \textbf{97.06} & \textbf{60.61} & \textbf{35.46} & \textbf{99.40} & \textbf{80.20} & \textbf{54.76} & 0.22          \\
FastVGGT  & 96.52          & 58.32          & 34.63          & 98.93          & 78.69          & 53.11          & {\ul 0.20}    \\ \midrule
FlashVGGT & {\ul 96.67}    & {\ul 58.44}    & {\ul 34.75}    & {\ul 99.11}    & {\ul 78.98}    & {\ul 53.78}    & \textbf{0.15} \\ \bottomrule
\end{tabular}%
}
\end{table}

\noindent  \textbf{Downsampling Alternatives.}~To justify our descriptor-based attention, we compared it against three alternative strategies with a similar computational budget and the same training scheme:
\textbf{(a) Input Downsampling:} Resizing input images before encoding.
\textbf{(b) Feature Downsampling:} Resizing DINO features immediately after the encoder.
\textbf{(c) Global Bottleneck:} Downsampling tokens before the global attention block and upsampling them afterward to restore resolution.
As shown in the table below, our descriptor-based attention outperforms all variants. The reason is that all other variants irreversibly destroy high-frequency details through downsampling, while our method maintains the encoder, frame attention, heads, and global queries at full resolution. By only compressing the global keys and values, we model inter-frame correspondences efficiently without sacrificing intra-frame details, and therefore achieve better reconstruction quality.

\begin{table}[h]
\centering
\caption{Comparison with downsampling alternatives.}
\resizebox{\columnwidth}{!}{%
\begin{tabular}{@{}l|cccccc@{}}
\toprule
Method                     & Abs Rel$\downarrow$ & Acc$\downarrow$ & Comp$\downarrow$ & APE$\downarrow$ & Time (s)      & Mem (GB)       \\ \midrule
(a) Input Downsampling     & 0.332               & 1.306           & 0.936            & 6.843           & \textbf{9.72} & \textbf{33.87} \\
(b) Feature Downsampling   & 0.312               & 0.954           & 0.923            & 5.560           & 10.87         & 34.23          \\
(c) Global Bottleneck      & {\ul 0.241}         & {\ul 0.810}     & {\ul 0.884}      & {\ul 4.630}     & 11.03         & 36.87          \\ \midrule
\textbf{Descriptor (Ours)} & \textbf{0.161}      & \textbf{0.512}  & \textbf{0.310}   & \textbf{2.733}  & {\ul 12.08}   & {\ul 37.21}    \\ \bottomrule
\end{tabular}%
}
\end{table}

\noindent \textbf{Inference Breakdown.}
We provide a detailed breakdown of the inference time across different components for 1,000-image sequences, measured in seconds.
\begin{table}[h]
\centering
\caption{Breakdown of the inference time.}
\resizebox{\columnwidth}{!}{%
\begin{tabular}{@{}l|ccccc@{}}
\toprule
          & Encoder       & Frame Blocks  & Global Blocks  & Reconstruction Heads & Total          \\ \midrule
VGGT      & 2.25          & 5.15          & 368.16         & 2.04                 & 377.60         \\
FlashVGGT & \textbf{2.24} & \textbf{5.15} & \textbf{25.93} & \textbf{2.04}        & \textbf{35.37} \\ \bottomrule
\end{tabular}%
}
\end{table}

\section{Discussions}\label{suppl:discussion}

\noindent \textbf{Limitations.}
While FlashVGGT achieves substantial efficiency improvements over VGGT~\cite{wang2025vggt}, several limitations merit discussion. First, our method exhibits a slight performance gap on short sequences, as evidenced in Tab.~\ref{tab:camera_results}. However, this gap diminishes with longer sequences where our architectural advantages become more pronounced. Second, similar to the original VGGT, our model's performance can degrade under challenging conditions involving large deformations or extreme lighting variations. This limitation, however, could potentially be mitigated through fine-tuning on domain-specific data without requiring architectural changes.
These aspects represent promising directions for future work in enhancing robustness and applicability.

\noindent \textbf{Design Space.}
While our framework achieves substantial efficiency improvements, its design space offers rich opportunities for future exploration. For instance, although our experiments in Tab.~\ref{tab:spatial_compression} demonstrate that a simple convolutional compressor underperforms interpolation, more sophisticated learnable architectures for token compression and selection remain unexplored. Investigating adaptive mechanisms that can dynamically adjust compression strategies based on input characteristics could potentially yield further performance gains.

\noindent \textbf{Integration with other architectures.}
While our primary evaluation is based on VGGT~\cite{wang2025vggt}, our descriptor-based attention mechanism is a general module that can be readily integrated into other architectures employing the alternating attention backbone~\cite{wang2025pi3, keetha2025mapanything, dens3r, MoRE2026}. This portability paves the way for developing more efficient variants of state-of-the-art models across various tasks, including metric reconstruction~\cite{keetha2025mapanything} and semantic 3D understanding~\cite{li2025iggt}.


%% file: tables/suppl/abl_aux.tex
\begin{table}[htp]
\centering
\caption{\textbf{Detailed ablations of auxiliary descriptor tokens}. Evaluated on 7Scenes~\cite{shotton2013sevenscene} with 500 input images.}
\label{tab:ablation-aux}
\resizebox{\columnwidth}{!}{%
\begin{tabular}{@{}l|ccccccc@{}}
\toprule
           & Abs Rel$\downarrow$ & CD$\downarrow$ & NC$\uparrow$   & APE$\downarrow$ & ARE$\downarrow$ & Time (s) & Mem (GB) \\ \midrule
w/o Cam tokens  & 0.066 & 2.849 & 64.01 & 3.908 & 8.115  & 12.99          & 33.40          \\
w/o First frame & 0.067 & 2.866 & 58.91 & 7.660 & 13.608 & 12.67          & 33.39          \\
w/o Key frames  & 0.067 & 2.859 & 63.68 & 4.183 & 8.123  & \textbf{12.04} & \textbf{33.33} \\ \midrule
Full model & \textbf{0.066}      & \textbf{2.748} & \textbf{64.12} & \textbf{3.904}  & \textbf{8.115}  & 12.99    & 33.40    \\ \bottomrule
\end{tabular}%
}
\end{table}

%% file: tables/suppl/abl_keyframe.tex
\begin{table}[htp]
\centering
\caption{\textbf{Comparison of different key-frame selection methods.}}
\label{tab:abl-keyframe}
\resizebox{\columnwidth}{!}{%
\begin{tabular}{@{}l|cccccll@{}}
\toprule
        & Abs Rel$\downarrow$     & CD$\downarrow$          & NC$\uparrow$            & APE$\downarrow$         & ARE$\downarrow$         & Time (s) & Mem (GB) \\ \midrule
Random     & 0.067 & 2.789 & 63.92 & 4.108 & 8.123 & 12.74 & 33.39 \\
Fix stride & 0.067 & 2.784 & 64.02 & 4.096 & 8.189 & \textbf{12.70} & \textbf{33.39} \\ \midrule
Cluster & \textbf{\textbf{0.066}} & \textbf{\textbf{2.748}} & \textbf{\textbf{64.12}} & \textbf{\textbf{3.904}} & \textbf{\textbf{8.115}} & 12.99    & 33.40    \\ \bottomrule
\end{tabular}%
}
\end{table}

%% file: figures/figure10/figure10.tex
\begin{figure}[thp]
    \centering
    \includegraphics[width=0.95\linewidth]{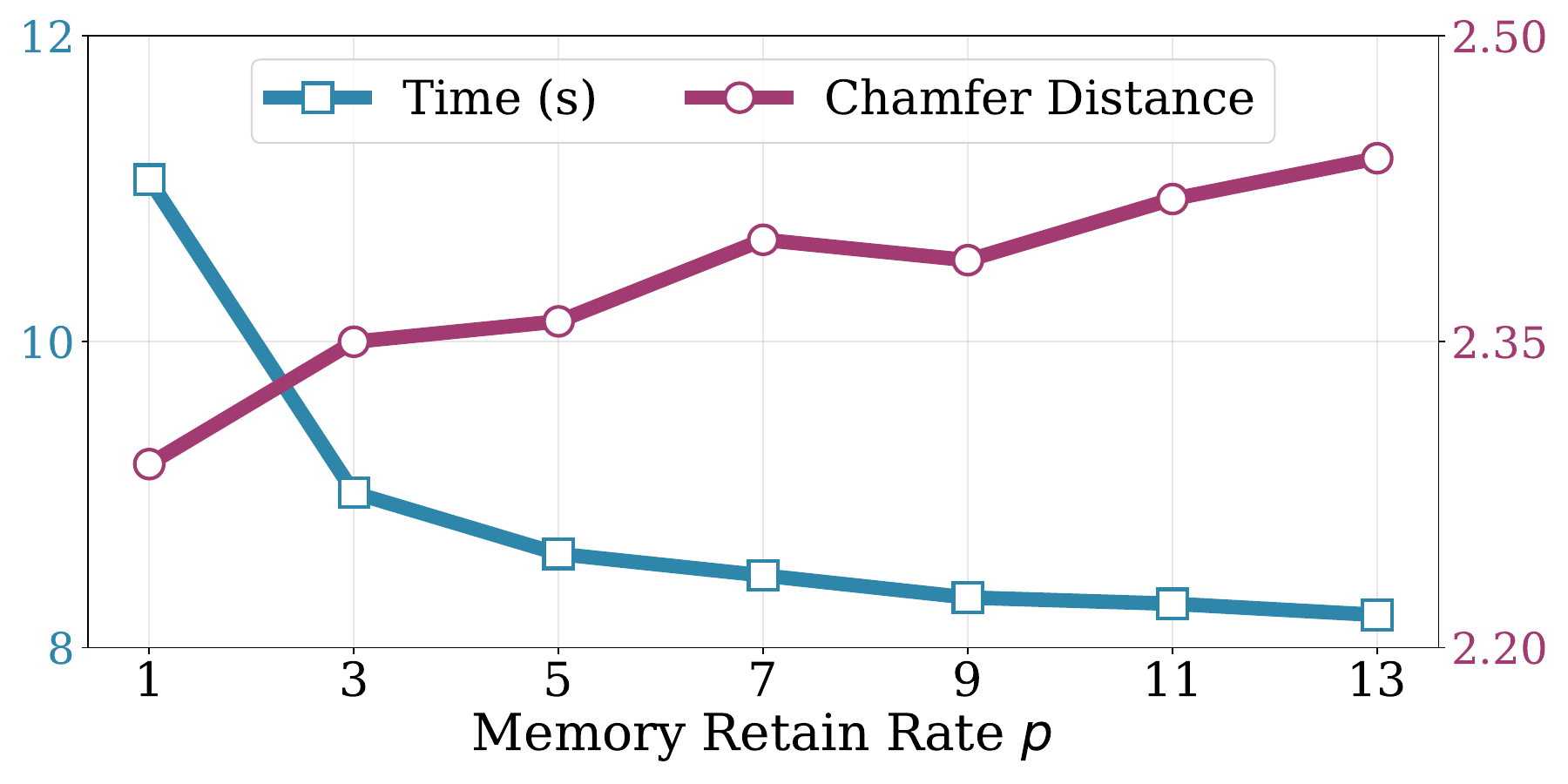}
    \caption{
    \textbf{The impact of different memory retain rate $p$.}
    The rate of 5 provides a balanced choice between quality and speed.
    }
    \label{fig:figure10}
\end{figure}

%% file: tables/suppl/abl_chunksize.tex
\begin{table}[htp]
\centering
\caption{\textbf{The impact of different chunk size in online setting. }}
\label{tab:chunk}
\resizebox{\columnwidth}{!}{%
\begin{tabular}{@{}c|ccccccc@{}}
\toprule
Chunk & Abs Rel$\downarrow$ & CD$\downarrow$ & NC$\uparrow$ & APE$\downarrow$ & ARE$\downarrow$ & \multicolumn{1}{l}{Time (s)} & \multicolumn{1}{l}{Mem (GB)} \\ \midrule
1   & 0.086 & 2.342 & 59.16 & 5.165 & 1.031 & 25.50 & 11.42 \\
10  & 0.087 & 2.391 & 59.74 & 5.042 & 0.980 & 14.58 & 13.33 \\
50  & 0.087 & 2.362 & 60.70 & 5.224 & 1.003 & 12.58 & 14.95 \\
100 & 0.087 & 2.330 & 60.97 & 5.173 & 1.034 & 10.90 & 16.98 \\ \bottomrule
\end{tabular}%
}
\end{table}

%% file: tables/suppl/resources.tex
\begin{table}[]
\centering
\caption{
\textbf{Comparison of resources consumption across different input images.}
'-' denotes model running out memory.
}
\label{tab:resources}
\resizebox{\columnwidth}{!}{%
\begin{tabular}{@{}c|l|cccccc@{}}
\toprule
\multicolumn{1}{l|}{}     & Methods  & 200   & 400   & 600    & 800    & 1000   & 1200 \\ \midrule
\multirow{3}{*}{Time (s)} & VGGT     & 17.01 & 61.82 & 137.84 & 245.47 & 386.07 & -    \\
                          & FastVGGT & 6.45  & 16.63 & 32.19  & 52.01  & 79.31  & -    \\ \cmidrule(l){2-8} 
 & \textbf{FlashVGGT} & \textbf{4.05}  & \textbf{9.84}  & \textbf{17.25} & \textbf{26.44} & \textbf{38.1}  & \textbf{51.25} \\ \midrule
\multirow{3}{*}{PFLOPs}   & VGGT     & 4.24  & 16.92 & 38.04  & 67.61  & 105.61 & -    \\
                          & FastVGGT & 0.59  & 2.23  & 5.19   & 9.20   & 14.34  & -    \\ \cmidrule(l){2-8} 
 & \textbf{FlashVGGT} & \textbf{0.29}  & \textbf{1.10}  & \textbf{2.43}  & \textbf{4.30}  & \textbf{6.70}  & \textbf{9.62}  \\ \midrule
\multirow{3}{*}{Mem (GB)} & VGGT     & 18.50 & 30.98 & 43.45  & 55.93  & 68.40  & -    \\
                          & FastVGGT & 19.34 & 32.66 & 45.97  & 59.29  & 72.60  & -    \\ \cmidrule(l){2-8} 
 & \textbf{FlashVGGT} & \textbf{16.97} & \textbf{27.92} & \textbf{38.83} & \textbf{49.76} & \textbf{60.68} & \textbf{71.61} \\ \bottomrule
\end{tabular}%
}
\end{table}

%% file: tables/suppl/perceiver.tex
\begin{table}[htp]
\centering
\caption{Comparsion between our compressed descriptor attention and Perceiver-style latent cross-attention.}
\label{tab:perceiver}
\resizebox{\columnwidth}{!}{%
\begin{tabular}{@{}l|cccccll@{}}
\toprule
                & Abs Rel$\downarrow$ & CD$\downarrow$ & NC$\uparrow$   & APE$\downarrow$ & ARE$\downarrow$ & Time (s)       & Mem (GB)       \\ \midrule
Perceiver-style & 0.097               & 5.645          & 34.02          & 14.573          & 12.564          & 13.56          & 34.56          \\ \midrule
\textbf{Ours}   & \textbf{0.066}      & \textbf{2.748} & \textbf{64.12} & \textbf{3.904}  & \textbf{8.115}  & \textbf{12.99} & \textbf{33.40} \\ \bottomrule
\end{tabular}%
}
\end{table}